# REVIEWING FID AND SID METRICS ON GENERATIVE ADVERSARIAL NETWORKS

Ricardo de Deijn, Aishwarya Batra, Brandon Koch, Naseef Mansoor and Hema Makkena

Department of Computer Information Science, Minnesota State University, Mankato, United States of America

## ABSTRACT

*The growth of generative adversarial network (GAN) models has increased the ability of image processing and provides numerous industries with the technology to produce realistic image transformations. However, with the field being recently established there are new evaluation metrics that can further this research. Previous research has shown the Fréchet Inception Distance (FID) to be an effective metric when testing these image-to-image GANs in real-world applications. Signed Inception Distance (SID), a founded metric in 2023, expands on FID by allowing unsigned distances. This paper usespublic datasetsthat consist of façades, cityscapes, and maps within Pix2Pix and CycleGAN models. After trainin , these models are evaluated on both inception distance metrics which measure the generating performance of the trained models. Our findings indicate that usage of the metric SID incorporates an efficient and effective metric to complement, or even exceed the ability shown using the FID for the image-to-image GANs.*

## 1. INTRODUCTION

Generative adversarial networks (GANs) have emerged as a powerful category of deep generative models and have been proven useful in the areas of image processing, computer generated graphics and computer vision applications [1]. Announced in [2], the authors refer to adversarial networks as the most exciting concept in Machine Learning in the past two decades [3]. Other approaches to generative modeling are based on optimization, but GANs utilize the concept of game theory and are implemented with two competing neural networks – a generator and a discriminator [2]. The task of the generator is to produce data that mimics the distribution of the original data whereas the discriminator network competes to identify the generator's output as real or fake. The two networks are locked in a game with each other where the generator generates an input that can trick the discriminator while the discriminator aims to improve its ability to identify the generated data from the real data. This adversarial tussle between the two networks leads to realistic newly generated data identical to the original data [2], [4]. Identical so much that the user of the data is not able to spot any difference.

Since their development, GANs have achieved remarkable success in generating realistic data across various domains [3]. A key advantage of these generative adversarial networks is their ability to learn feature representations without manual labeling of the training data [4]. GANs have been implemented in machine learning and artificial intelligence for purposes such as malware detection, speech and language processing and chess game programs [3]. They are also the driving force behind image-to-image translation tasks, which has the goal of translating one image to a different image [5]. This task is important for several objectives, to give a few examples: it can be used for data augmentation to create more input datasets for computer vision tasks; It can also be used for photo editing and photo resolution improvement, like used in photo editing applications [6].

*Image-to-Image translation GANs:* When GANs are used as a way of translating one image to a different image, it is called Image-to-Image (I2I) translation GANs [5]. There are a wide variety of Image-to-Image translators, such as:

- Supervised Image-to-Image Translation – This type of GAN is trained on both an input and target image dataset. The images within these datasets are paired with one another in such a way that the model can train to which domain an input needs to be translated to [5].
- Unsupervised Image-to-Image Translation – Most of the time, paired image datasets are what supervised image-to-image translation GANs require. However, these datasets are not widely available, and it may take a lot of time to pair up images from one dataset exactly with a similar image of another dataset. Therefore, unsupervised Image-to-Image translation has acquired more attention recently. Unsupervised Image-to-Image translation makes use of two large datasets to train the GAN, but without specifying which images of one dataset are in direct relationship with an image for the other dataset [5].
- Semi-Supervised Image-to-Image Translation – When there is limited paired data between the input and target datasets, it is possible to train an Image-to-Image translation GAN in a semi -supervised way. It uses the images that are labeled to uncover the content and style domain but uses the unpaired images to generalize the connection between the real images and translated images [5], [7].
- Few-shot Image-to-Image Translation – Humans have the ability to train their brain on just one or a couple of images rather than requiring an entire dataset of images to uncover the content of the image. This concept is implemented in few-shot GANs to improve results on small, incomplete datasets. Small, incomplete datasets are common as it is expensive to acquire these complex datasets. Additional experiences and knowledge using meta data can be used to improve output results. This variety of GAN necessitates that the limited 'few-shots' are matched between the input and output image datasets, analogous to the paired datasets utilized in supervised image-to-image translation [5].

These GANs can be used for various goals. These goals entail domain adaptation, which transforms styles from an image to a different domain style, or style transfer, which keeps most of the input content while changing its style aspects [5], [8], [9]. However, current I2I GAN models still pose problems of mode collapse, instability, and lack of diversity [1]. This lack of diversity can be addressed by evaluating the GANs on metrics that measure the diversity between the real and generated images. A metric like Fréchet Inception Distance (FID) tracks this diversity by measuring the inception distance between the real and generated data feature distributions [10],whereas the metric of Signed Inception Distance (SID) allows improvements by accepting an unsigned distance score [8]. Within this paper, we will compare the SID and FID metrics against two GANs trained on three datasets and determine how both metrics can help specify which GAN is better at recreating a certain domain.

The remainder of the paper is organized as follows - Section 2 provides a brief introduction about Pix2Pix and CycleGAN highlighting their internal working. Section 3 focuses on the methodology used in our research detailing the datasets and models used for evaluation. In section 4, the experimental results and comparative analysis are provided based on FID and SID explaining the criteria used for analysis. Section 5 outlines the conclusion of the research demonstrating the difference in score between SID and FID, and which type of image data each GAN performs better on. Finally, in section 6 we conclude by discussing the challenges and scope for future research.

## 2. RELEVANT WORK

Pix2Pix, an early Image-to-Image GAN, pioneered high-quality image generation and influenced subsequent models. Isola et al. introduced, in [11], automatic image-to-image translation, teaching the model both image mapping and the loss function for versatile translation. It employs a U-Net generator with skip connections, avoiding low-level information bottlenecks, as is visualized in Figure 1. Pix2Pix excels with small datasets, and it relies on paired images for supervision and underutilizes noise, addressed with dropout. This enhances Pix2Pix' adaptability beyond its training data [6].

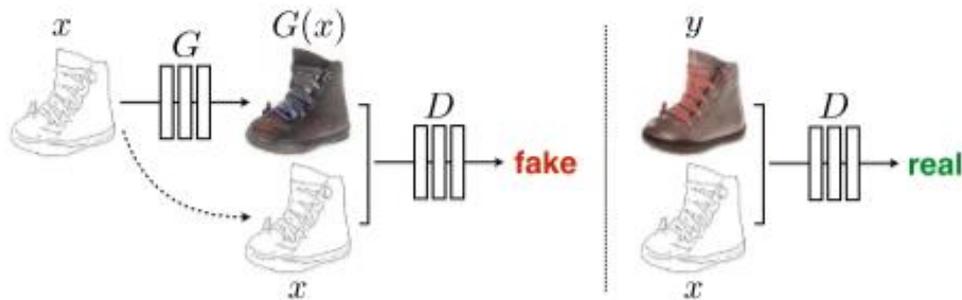

Figure 1. Schema on how the Pix2Pix GAN trains on how to translate edges to real looking shoes. The discriminator identifies the difference between ground truths and generated images, and the generated learns to mimic the ground truth target images. Adapted from [11].

However, Pix2Pix GAN lacks cycle consistency, which would improve consistency verification for image translations results between two domains. For this reason, CycleGAN [12] is introduced, as an extension of Pix2Pix with cycle consistency. It tackles image-to-image translation without paired data by matching the distribution of generated images to real ones. It employs a 70 x 70 PatchGAN [12] discriminator and a generator with 6-9 residual blocks and fractionally-strided convolutions. CycleGAN introduces adversarial and cycle consistency losses to maintain image set characteristics across domains as visualized in Figure 2. Unlike Pix2Pix, it uses an autoencoder structure, lacks skip connections, and doesn't employ a conditional GAN [6], [9], [12].

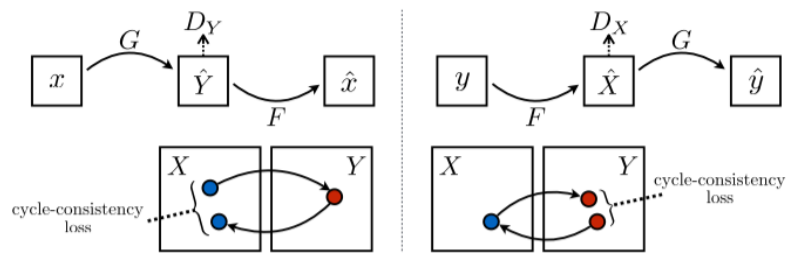

Figure 2. The CycleGAN two mapping function, including the forward cycle-consistency loss and backward cycle-consistency loss. Adapted from [12].

## 3. METHODOLOGY

### 3.1. Inception Distance Metrics

FID is a popular metric used to evaluate the quality of images generated by GANs. It has become popular, as there is a high correlation with FID scores and interpretation of the human eye [10]. It measures the distance between the multivariate Gaussian distributions of the generated image data set and the real data that the GAN is trying to replicate (ground truth) with a formula equivalent to the 2-Wasserstein formula [10], [13].

$$FID(\mu_r, \Sigma_r, \mu_g, \Sigma_g) = \left\| \mu_r - \mu_g \right\|_2^2 + Tr(\Sigma_r + \Sigma_g - 2(\Sigma_r \Sigma_g)^{\frac{1}{2}}), \quad (1)$$

in which $(\mu_r, \Sigma_r)$ and $(\mu_g, \Sigma_g)$ stand for the sample mean and the covariance of the feature approximation of the ground truth and generated data. $Tr(\cdot)$ indicates the matrix trace [10]. InceptionV3 model, which is pre-trained on the ImageNet dataset, is used to extract the feature vectors for each image in a dataset, resulting in a multivariate Gaussian distribution of features over the dataset. For practical purposes, only mean and covariance are considered to model both real data and generated data as gaussian distributions. It has been noticed that FID judges the generated images as well as how a human would judge the images as it gives an overall evaluation of image quality and diversity [8], [14], while at the same time, it is quite simple and fast to get to a score. All this results in that FID is a quite common metric in the GAN development sphere. FID scores on an interval of 0 to infinity. When an FID score is low, the quality and diversity ofthe generated image dataset are greater [10]. Figure 3 visualizes the process on how a Fréchet Inception Distance gets calculated between a ground truth and generate image dataset.

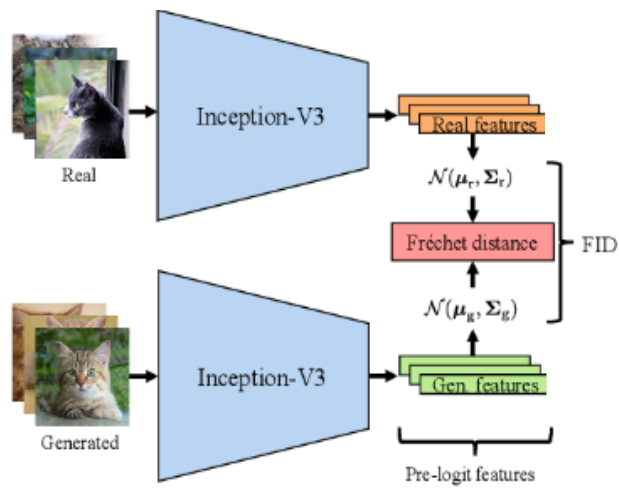

Figure 3. The FID gets calculated by first passing the ground truth and generated images through the Inception-V3 classifier network. This will result in two sets of feature vectors containing important aspects of the images. Afterwards the distributions of these features are approximated with multivariate Gaussians and the result is the Fréchet distance between the two Gaussians. Adapted from [10].

SID metric expands on the fundamental ideas of FID and incorporates enhancements from Precision-Recall analysis. The added complexity of having an unsigned distance score makes it a feasible alternative to FID [8], [14]. SID is a modern statistic that aims to capture the variation and diversity found within artificially generated images using formula [8]

$$SID_{m,r}(\mu_p||\mu_q) = \frac{1}{NM_x} \sum_{\substack{l=1 \\ x_l \in C'_{q,r}}}^{M_x} \left( \sum_{\substack{j=1 \\ c_j \sim \mu_q}}^{N} \Phi(x_l, \psi(c_j)) - \sum_{\substack{i=1 \\ \tilde{c}_i \sim \mu_p}}^{N} \Phi(x_l, \psi(\tilde{c}_i)) \right), \qquad (2)$$

where N is batches of samples, $C'_{q,r}$ is a hypercube of side r centered on the transformative distribution $\psi(c_j)$, $M_x$ is number of test points uniformly sampled within the hypercube $C'_{q,r}$. $\Phi(x_l, \psi(c_j))$ is a kernel function used to measure the interaction between two points in the feature space, influenced by their distance and other factors [8].

It might be especially helpful if we want to evaluate the various content and stylistic elements included in the generated images. Therefore, SID compliments FID for assessing GAN performance by evaluating on an interval of -infinity to infinity. Since SID concentrates on identifying the diversity as well as variation within generated images, whereas FID just assesses how closely generated images resemble real photos.

### 3.2. Data

In this work we have used three different publicly available datasets: Facades [15], Maps [11], and Cityscapes [16] to evaluate the Pix2Pix and CycleGAN. We discuss these datasets in the following subsections.

In this research, building façades data, created by the Center for Machine Perception as named in [15], was used. This dataset is a paired dataset that includes images of different facades or fronts of buildings around the world paired to the outline of their unique characteristics as Figure 4 shows. Further, the unique architecture on these buildings brings new perspectives in the 606 images that are provided. Some of the unique characteristics of the buildings include different shapes and sizes of the doors, windows, and pillars. Additionally, each building has a different background and some even include diverse types of balconies and molding.

Figure 4, shown below is a small glimpse into the overall dataset. Just in this snippet alone, generative models represent the accents of each building quite differently. The dataset serves as a valuable resource for image-to-image translation tasks due to its accessibility, the distinctiveness of each image, and the level of difficulty it presents for GAN models. The challenge stems from the previously mentioned diversity of the building's doors, windows, pillars, roofs, etc.

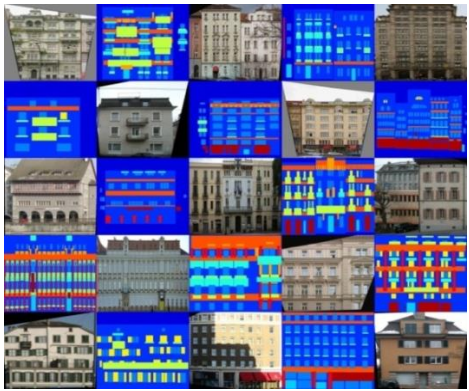

Figure 4. Sample images from Facades dataset representing examples of pairs that can be found in the dataset. Adapted from [15].

A second dataset that is utilized is a dataset about city landscapes. The dataset was proposed in [16] and made public at the Conference on Computer Vision and Pattern Recognition (CVPR) 2016 and available on the Berkeley AI Research Group website. This cityscape dataset is a paired dataset that includes images of frontal views of the city taken from within a car and a semantic view of the outline of these cityscapes, as is shown in Figure 5. These images can be used when the GAN properly works to transform images to semantic images and the other way around, so it can be used for semantic object labeling or creating cityscapes from outlines. The dataset utilizesa lot of textures and colors, which as a result makes it hard to generate new cityscapes or new semantic images as there are so many possibilities.

The data includes 2975 train images and 500 train images, making the total 3475 images. The semantic outlines of the cityscapes include outlines of the road, poles on the side of the road, houses, trees, traffic lights, and pedestrians with each their own colors.

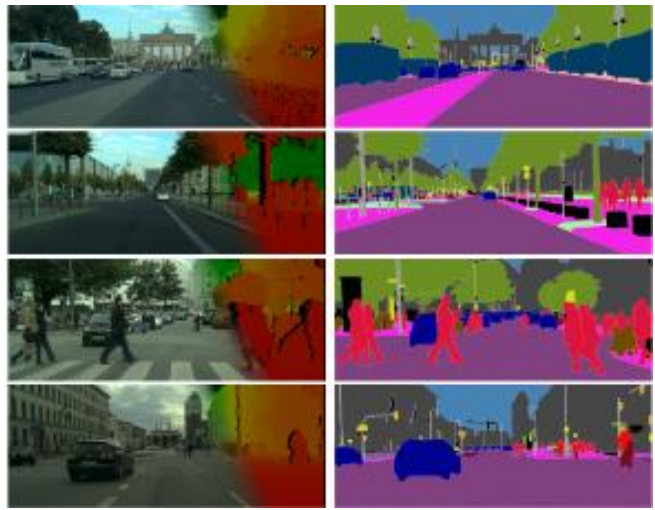

Figure 5. Sample images from the Cityscape dataset representing examples of pairs that can be found in the dataset. Adapted from [16].

As a third dataset, Satellite Map dataset from [11] is used. Like previous datasets, this dataset is also paired and has map images of various locations as well as their satellite images. What makes this dataset interesting is the domain difference between the two images. Even after having a domain difference, what makes this dataset suitable for use in pix2pix GAN is correlation between the two images at pixel level. It has 1096 images in the train dataset and 1098 in the validation set. Below, in Figure 6, is a small snippet shown from the satellite-map dataset which shows satellite images and map images [17].

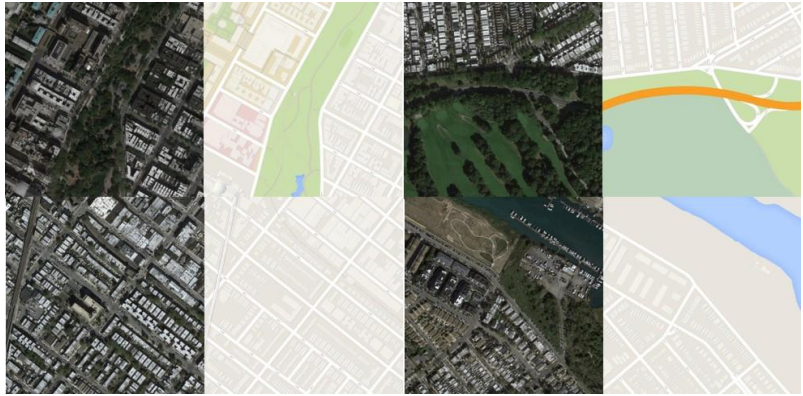

Figure 6. Sample images from Maps dataset representing examples of pairs that can be found in the dataset. Adapted from [17].

### 3.3. Models

This paper will compare results between the FID and the recently proposed metric SID on two older Image-to-Image translation GANs. These GANs are Pix2Pix [11] and CycleGAN [12] and have been selected for the fact that they have been important models in the development of GANs. Pix2Pix was the GAN that introduced Image-to-Image translation [11]. CycleGAN is a continuation on Pix2Pix, improving the consistency between input and generated images [12].

These models were originally evaluated without using either of the inception distance metrics to measure the diversity between the ground truth and generated image dataset distributions. Therefore, it is an interesting selection to test both inception distance metrics on. This was because both FID and SID were not around when these models were first proposed. To test how well inception distances metrics can be used to evaluate models, we will use these two former state-of-the-art GAN models to see how well FID and SID perform on all GANs. All our code bases can be found on [18].

## 4. RESULTS

### 4.1. Experimental Results

To stem our research from the creators of FID and SID, we used the FID PyTorch implementation from [19] and the clean-SID GitHub repository mentioned in [8].

Due to the complexity of the Façade dataset, we used 100 epochs for Pix2Pix. By training the GAN on the training data 100 times, the generator gets more successful at recognizing the key features from the input distribution. This results in the generated images and the ground truth becoming more similar. For CycleGAN, the model iterated over the training data 20 times. In Figure 7, the generated image of the bottom row was trained from scratch using 100 epochs in the Pix2Pix GAN, resulting in high similarities between the generated and ground truth images.

118   Computer Science & Information Technology (CS & IT)

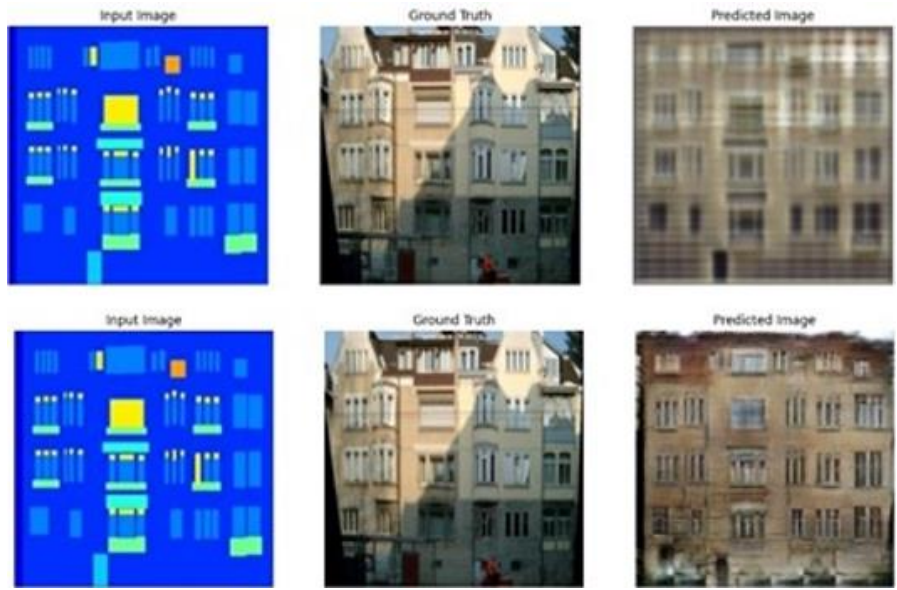

Figure 7. Output generated by Pix2Pix GAN after training for 1 epoch (top) and 100 epochs(bottom) on facades dataset.

If we look at the top row of figure 8, we can see that after 1 epoch the generated image is quite primitive. The street, buildings and vehicles are barely discernible with blurred lines. It lacks sophistication and is an abstract representation of the input image. After 100 epochs, the details have improved. The same street, vehicle and buildings have sharper detail and offer a more realistic view of the input image than before.

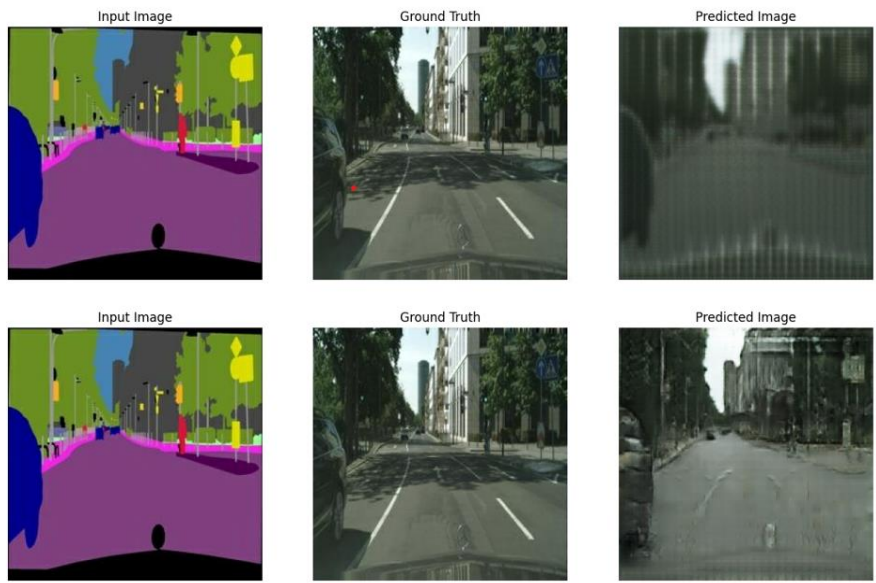

Figure 5. Output generated by Pix2Pix GAN after training for 1 epoch (top) and 100 epochs (bottom) on cityscapes dataset.

The top row of figure 9 is a very rudimentary structure with just outlines of objects. It is impossible to make out what the generated image is. After 100 epochs, details have considerably improved. The image is clearer with significant enhancement in sharpness of the image.

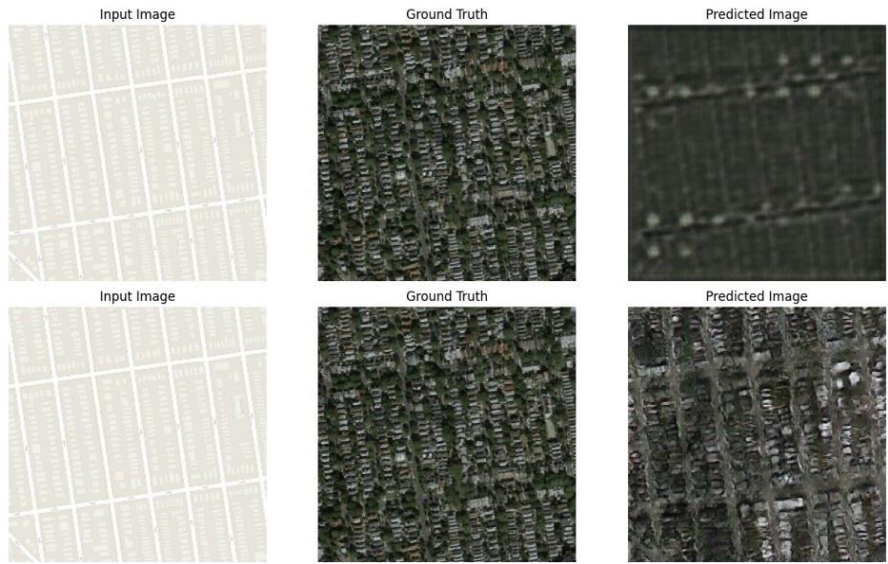

Figure 6. Output generated by Pix2Pix GAN after training for 1 epoch (top) and 100 epochs (bottom) on maps dataset.

The top row of Figure 10 shows a generated image of a building from the Façade dataset trained on the GAN model CycleGAN. This image is blurry to the point that it does not look like a building. On the other hand, the bottom row of Figure 10, which is the generated image after 8 training iterations of the training data, resembles more closely to a building, although not perfect.

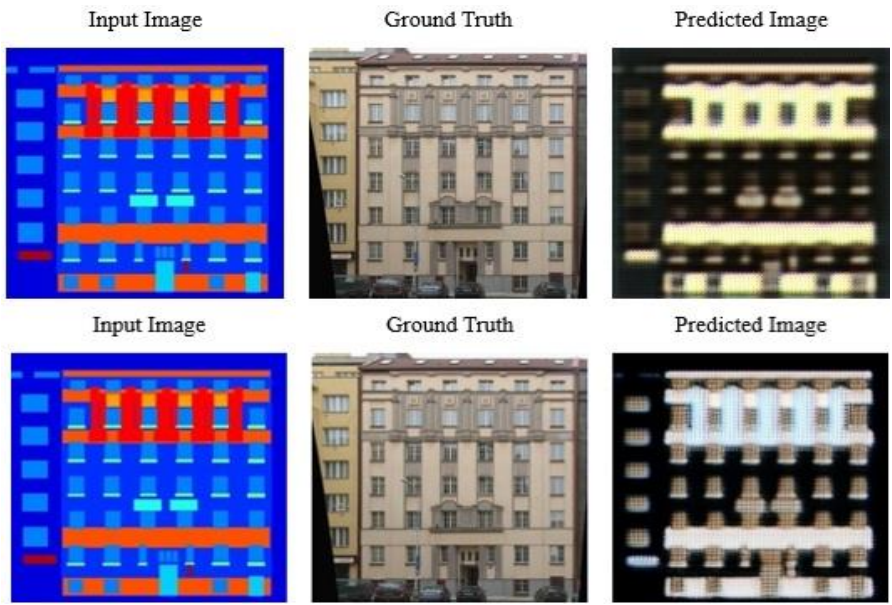

Figure 7. Validation output generated by CycleGAN GAN after training for: 1 epoch (top), and 8 epochs (bottom) on facades dataset.

After training the cityscape dataset for 1 and 8 epochs, the CycleGAN GAN model is able to catch some concepts from the ground truth dataset as can be seen in Figure 11. Compared to the facades dataset, a similar trend is visible. The model cannot catch specific details from the ground truth as well as Pix2Pix can do after 100 epochs in Figure 8. However, after 8 epochs, it is possible to spot that the pictures are getting sharper, similar to what happened in Figure 10.

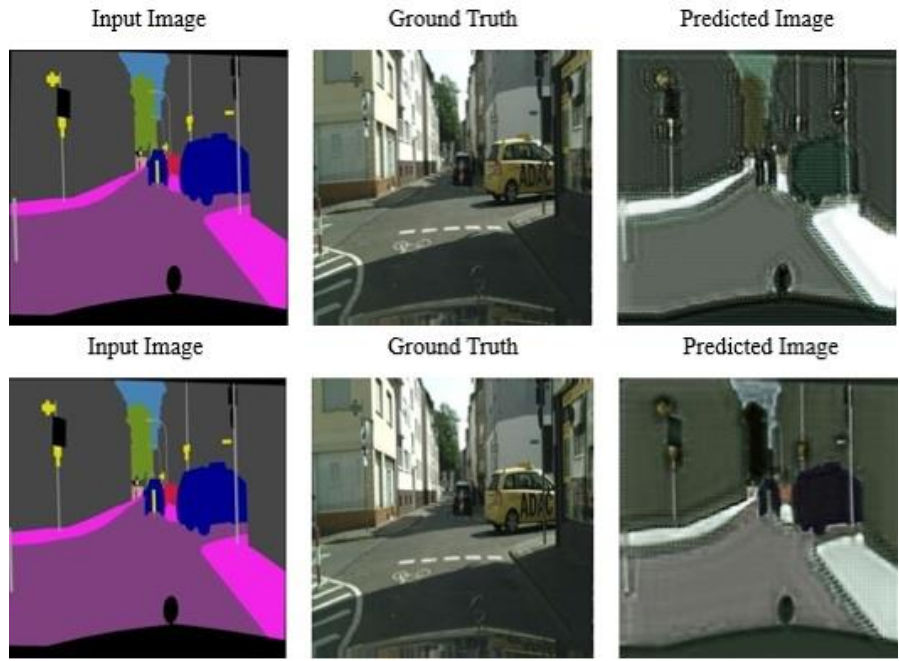

Figure 8. Validation output generated by CycleGAN GAN after training for: 1 epoch (top), and 8 epochs (bottom) on cityscapes dataset.

After processing the maps dataset with CycleGAN, comparable results were found after epoch 0 and epoch 7, as shown in Figure 12. However, this time the goal was the other way around, to process satellite images to a maps representation. CycleGAN provides more details and sharper results after 8 epochs but does not look similar to a maps representation yet, with an exception on its color usage in the bottom row.

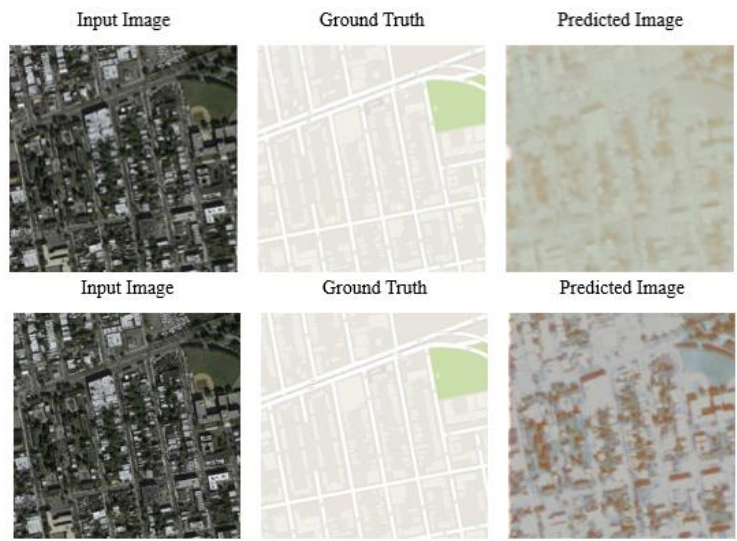

Figure 9. Test output generated by CycleGAN GAN after training for: 1 epoch (top), and 8 epochs (bottom) on cityscapes dataset.

As training progresses and the number of epochs increase, there is a significant increase in realism in the generated images. This can be correlated with the improvement in inception distance scores. The FID and SID scores, which are critical indicators of image quality and diversity, are expected to show significantly better scores for models with a higher epoch count like Pix2Pix compared to CycleGAN with a lower epoch count.

### 4.2. Comparison Results Based on SID and FID

After training Pix2Pix on the Facades dataset [15], an FID score of 162.1522 is acquired. On these same results an SID score of 120.5520. Datasets with lower diversity than the target have a negative SID score and when the target distribution has a lower diversity than the input dataset, we have a positive SID score. When the two distributions have an equal diversity, the SID score is close to 0 or exactly 0. The Pix2Pix model's SID score 120, which means that the ground truth distribution is more diverse than the generated dataset.

After training CycleGAN on the Facades dataset [15] for 8 epochs, an FID score of 316.42875 got collected. This time the SID score is further away from zero than FID, which has a 14782.1975 score. This time SID gives more weight to the fact that the ground truth distribution is way more diverse than the generated image dataset, which is a result of the limited number of epochs that it has ran on.

The Pix2Pix and CycleGAN have also been trained on the Cityscapes dataset [16]. Each again for 100 and 8 epochs. As can be seen from the images trained with the Pix2Pix in Figure 8, the Pix2Pix model seems to be able to capture the meaning of the ground truths better after 100 epochs than the CycleGAN is able to do after 8 epochs as shown in Figure 11. This is also shown in the SID and FID scores for each of these models.

The Pix2Pix model achieves an FID score of 174. 4255 and SID score of 4420.7079 on the cityscapes dataset. This all while the CycleGAN model is only able to achieve an FID score of 227.7994 and an SID score of 8667.2221.

At last, both Pix2Pix and CycleGAN are trained on the maps dataset [11] that translates satellite images to images similar to Google Maps. Once again Pix2Pix scores lower on both FID and SIDwith an FID score of 213.1165 and SID score of 3184.3623 against the FID score for CycleGAN at 253.0294 and SID score of 9316.9300.

Table 1. FID and SID scores for Pix2Pix and CycleGAN GAN models on the Façade [15], Cityscape [16], and Maps [11] dataset.

|  | Façade | | Cityscape | | Maps | |
|---|---|---|---|---|---|---|
|  | FID | SID | FID | SID | FID | SID |
| Pix2Pix *after 1 epoch* | 381.7384 | 13132.2067 | 397.3368 | 13128.3172 | 232.7371 | 6025.8421 |
| Pix2Pix *after 8 epochs* | 252.3915 | 600.9891 | 397.3740 | 13159.2410 | 205.0118 | 2845.2790 |
| Pix2Pix *after 100 epochs* | 162.1522 | 120.5520 | 174.4255 | 4420.7079 | 213.1165 | 3184.3623 |
| CycleGAN *after 1 epoch* | 347.3554 | 13671.4148 | 229.4415 | 5805.8483 | 299.1129 | 11242.5304 |
| CycleGAN *after 8 epochs* | 316.4288 | 14782.1975 | 227.7994 | 8667.2221 | 253.0294 | 9316.9300 |

All by all, it has become clear from both the generated images in section 4.1, as the FID and SID scores in Table 1 that Pix2Pix generates pictures of higher quality than the same pictures generated by the CycleGAN. However, it is also important to note that Pix2Pix has been trained on a higher epoch count than CycleGAN. Besides that, it is possible to see that dataset cityscape and maps, which contain more training and test examples than façade, score a higher SID and FID score for Pix2Pix. This could mean that the train and test set are a substantial influence on the FID and even more on the SID score. The SID scores give more weight to datasets with greater diversity than FID does, resulting in a high SID score. On the other side is it also possible that the FID and SID score lower on the CycleGAN model for the datasets that have more training and test examples. Meaning that CycleGAN is still severely undertrained for facades dataset. All SID and FID scores for CycleGAN are relatively high, which means that CycleGAN is not able to stabilize itself on these datasets after 8 epochs.

## 5. CONCLUSION

We can conclude that FID and SID are closely related to each other, showing that SID is a feasible alternative to the more commonly used evaluation metric FID. However, SID is able to give a more accurate representation on the distance between the ground truth distribution and the newly generated data distributions on the same epoch level as it is also able to provide a negative distance score in the instance the generated data distribution would be more diverse than the ground truth distribution. As a result, SID can provide more accurate weights to the distance between the two dataset distributions and shows a higher distance in case the generated dataset is far away from the ground truth image dataset.

We are also able to see that the size of the dataset is a heavy influence of the performance of a GAN. Both Pix2Pix and CycleGAN perform better on the cityscape image dataset, which is about 5 times the size as the facades dataset, based on their FID and SID scores. It can be said that Pix2Pix in general scored better than CycleGAN as well. This proves that the number of epochs is particularly important, and that a small number of epochs result in a very instable generated dataset, which is highlighted through the remarkably high SID scores of both CycleGANs.

## 6. DISCUSSION

Current literature showed us that FID is currently a go-to metric for GAN development thanks to its excellent correlation with human judgement. However, at the same time, we saw that the SID metric proposed earlier in 2023 with SpiderGAN in [8] brought the possibility of a signed inception distance between two data distributions. This results in the metric can now also measure when a generated dataset is more diverse than its ground truth. We wanted to try out if this would also been the case for earlier developed GANs. Although we obtained, in this research, that SID is a good addition to the FID metric in finding the diversity distance between the ground truth and generated data distributions, there is no single metric that can find the most perfect results. A great approach would be to use a combination of multiple metrics to evaluate the diversity and performance of newly generated images of freshly trained GANs.

### 6.1. Limitations

Originally, we were trying to use MNIST and Fashion MNIST for training our pix2pix model. The MNIST contains grayscale images of digits and Fashion MNIST contains grayscale images of clothing items. We tried to map digit to fashions items, for example 1 for trousers, 2 for shoes which we were unsuccessful at because there is domain mismatch which pix2pix is not good at handling. The model was unable to learn features from MNIST to directly map to fashion items in fashion MNIST. To overcome this challenge, we chose the facades, cityscapes, and maps datasets, which have a more meaningful relationship between the input and output. One of the primary limitations of our research was reliance on these small datasets which impacted the GAN training performance. While the current datasets were able to show that SID is a great addition to FID, the current GANs were unable to capture the diversity and variability well in these smaller datasets as it would compared to larger and more diverse datasets.

Besides the dataset problem, we also faced problems with the SID metric. We acquired the Clean-SID metric from the GitHub repository as mentioned in [8], but when we tried to run it, we found out that there were a lot of bugs present in this research paper's code. As this was the only repository available on the Signed Inception Distance, we were forced to debug the code to be able to calculate the distance score for each result through this metric. The debugged version of SID can be found on [20].

### 6.2. Future Research

By using the two inception distances our research has heavily looked at the diversity of the images in the Façade dataset. This means that the lack of diversity or quality of the images was examined. There is exciting potential to further this research. We have only investigated the diversity of images, however, the consistency of colors between real and fake images was not. Furthermore, human inspection of the images between the two metrics used was not considered. The choice of what to investigate will always depend on what the end goal is. With the recent implementation of Signed Inception Distance (SID) into image-to-image translation, further research is vitally important for continued improvements to this field.

Furthermore, our research focused on Pix2Pix and CycleGAN using three datasets. More research should be explored with implementation of FID and SID using other image-to-image GANs, such as MUNIT, and using various larger datasets. These datasets should vary in size, color, and overall complexity. Again, the dataset will vary depending on the overall application of the user. It would therefore also be interesting to see how FID and SID would score if datasets with different domains were mixed in both the ground truth as generated image datasets. FID and SID should score very high, since the diversity between both would be different. But if the images from different domains are mixed on a similar scale for both the generated as ground truth images, a SID and FID score could theoretically be achieved that is as good as if it were from one domain.